\pdfoutput=1

\PassOptionsToPackage{sort}{natbib}

\documentclass[11pt]{article}

\usepackage{ACL2023}

\usepackage{times}
\usepackage{latexsym}
\usepackage{multirow}
\usepackage[T1]{fontenc}

\usepackage[utf8]{inputenc}
\usepackage{enumitem}
\usepackage{microtype}

\usepackage{inconsolata}
\usepackage[most]{tcolorbox} 
\usepackage{mdframed}
\usepackage{xcolor} 
\usepackage{booktabs}
\usepackage{graphicx}

\usepackage[skip=5pt]{caption}

\newcommand{\header}[1]{\vspace*{1mm}\noindent\textbf{#1}.}

\author{Amin Abolghasemi\textsuperscript{1}, Zhaochun Ren\textsuperscript{1}, Arian Askari\textsuperscript{1}, Mohammad Aliannejadi\textsuperscript{2} \\  \textbf{Maarten de Rijke\textsuperscript{2}}, \textbf{Suzan Verberne\textsuperscript{1}} \\
  \textsuperscript{1}Leiden University, Netherlands \\
  \textsuperscript{2}University of Amsterdam, Netherlands \\
  \texttt{\{m.a.abolghasemi, z.ren, a.askari, s.verberne\}@liacs.leidenuniv.nl} \\
  \texttt{\{m.aliannejadi, m.derijke\}@uva.nl}
}

\title{CAUSE: Counterfactual Assessment of User Satisfaction Estimation\\ in Task-Oriented Dialogue Systems}

\begin{document}

\maketitle

\begin{abstract}
An important unexplored aspect in previous work on user satisfaction estimation for Task-Oriented Dialogue (TOD) systems is their evaluation in terms of robustness for the identification of user dissatisfaction: current benchmarks for user satisfaction estimation in TOD systems are highly skewed towards dialogues for which the user is satisfied. 
The effect of having a more balanced set of satisfaction labels on performance is unknown.
However, balancing the data with more dissatisfactory dialogue samples requires further data collection and human annotation, which is costly and time-consuming.
In this work, we leverage large language models (LLMs) and unlock their ability to generate satisfaction-focused counterfactual dialogues to augment the set of original dialogues of a test collection.  
We gather human annotations to ensure the reliability of the generated samples. 
We evaluate two open-source LLMs as user satisfaction estimators on our augmented collection against state-of-the-art fine-tuned models. 
Our experiments show that when used as few-shot user satisfaction estimators, open-source LLMs show higher robustness to the increase in the number of dissatisfaction labels in the test collection than the fine-tuned state-of-the-art models.
Our results shed light on the need for data augmentation approaches for user satisfaction estimation in TOD systems. 
We release our aligned counterfactual dialogues, which are curated by human annotation, to facilitate further research on this topic.
\end{abstract}
\section{Introduction}
\label{sec:intro}
Task-oriented dialogue (TOD) systems help users complete specific tasks, e.g., booking a hotel or restaurant, through conversations~\cite{feng2022topic,sen2023aaai,wang2022task,zeng2023futuretod}.
\begin{figure}[ht]
    \centering  \includegraphics[width=\columnwidth]{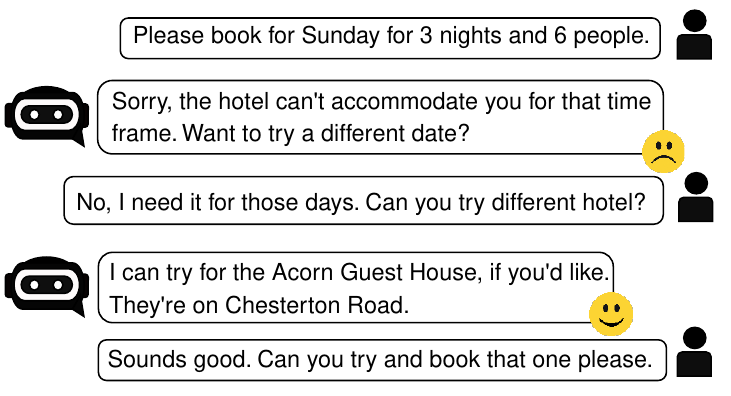}
    \caption{Example dialogue (snippet) between the user and the system from the MultiWOZ benchmark.}
    \label{fig:snippet}
\end{figure}
User satisfaction estimation (USE) is a key task in TOD systems, aiming to measure the extent to which users are satisfied with the dialogue they are having with the system (see Figure~\ref{fig:snippet}). 
USE has various applications as it can be viewed as a continuous approximation of human feedback for the quality of the dialogue. 
Such feedback enables human intervention for users who are having a dissatisfactory dialogue with the system. Furthermore, it serves as a scalable method for the automatic evaluation of dialogue systems and helps identify and optimize a dialogue system's shortcomings~\cite{ye2023modeling,song2023speaker}. 
Prior work has studied user satisfaction estimation in TOD systems~\cite{sun2021simulating,deng2022user,hu2023unblock, ye2023modeling} based on the user satisfaction simulation (USS) benchmark, which consists of several datasets annotated with user satisfaction labels by \citet{sun2021simulating}. However, the robustness of user satisfaction estimators for the identification of user dissatisfaction is an unexplored aspect in these works as most of the datasets are highly skewed towards the dialogues for which the user is satisfied. 

Put another way, the impact of a more balanced set of satisfaction labels on the performance of the USE models remains unknown. Nevertheless, balancing the data with more dissatisfactory dialogue samples demands further dialogue collection and human annotation which is costly and time-consuming.

To begin to address the issues raised above, we aim to expand the current imbalanced benchmarks of TOD systems with more dissatisfactory dialogues. To this aim, we leverage large language models (LLMs) and unlock their ability to generate counterfactual task-oriented dialogue samples.
We use counterfactual utterance generation to generate counterpart dialogue samples with an opposite satisfaction score for a given input dialogue sample, thereby increasing the number of dissatisfaction-labeled samples in the test collections.
Following the definition of user satisfaction and the annotation guidelines from the original work in which MultiWOZ~\cite{eric2020multiwoz} and SGD~\cite{rastogi2020towards} were annotated for user satisfaction levels,\footnote{We contacted the authors of \citep{sun2021simulating} in which the datasets were originally annotated with satisfaction scores.} we conduct human annotation on the counterfactual dialogues to ensure the quality and reliability of the generated utterances. By doing so, we introduce two augmented versions of the test collections for Multi\-WOZ and SGD benchmarks.

We focus on \emph{binary} satisfaction levels, i.e., dissatisfaction and satisfaction. We argue that (i) binary labels reduce the subjectivity of annotators in labeling the dialogue, and (ii) binary satisfaction could be more relevant in some TOD system contexts, since in real-world use cases, e.g., post-hoc analysis of dialogue systems, one might only look for identification of the cases where the user is dissatisfied with the dialogue and discard the cases where the dialogue proceeds smoothly and normally. 
In other words, for our purposes classifying whether a dialogue is \emph{dissatisfactory} or not is of more importance than classifying a \emph{normal} (rating 3 in a five-point scale satisfaction levels) or \emph{satisfying} (rate 4) from a \emph{very satisfying} dialogue (rate 5).
Table~\ref{tab:original_data_statistics} shows both the five-point scale and the binary-level mapping of the MultiWOZ and SGD datasets used by \citet{sun2021simulating}. 
As Table~\ref{tab:original_data_statistics} indicates, the current evaluation test collections for user satisfaction estimation in TOD systems are highly imbalanced towards the \emph{normal} satisfaction label (3). In the binary-level satisfaction setting, this imbalance results in most dialogue samples being annotated with \emph{satisfaction} labels, while the remaining samples are labeled as \emph{dissatisfaction}. 

\begin{table}[]
    \centering
    \renewcommand{\arraystretch}{1}
    \scalebox{1}{
    \begin{tabular}{l rr}
    \toprule
        Rating & MultiWOZ & SGD 
        \\
        \midrule
         1 & 12 & 5 
         \\
         2 & 725 &  769 
         \\
         3 &  11,141 & 11,515 
         \\
         4 & 669 & 1,494 
         \\
         5 &  6 & 50
         \\
         \midrule
         Dissatisfaction & 737  & 774
         \\
         Satisfaction & 11,816 & 13,059
         \\
         \bottomrule
    \end{tabular}
    } 
    \caption{Data statistics of MultiWOZ and SGD on five-point and two-point satisfaction scales.}
    \label{tab:original_data_statistics} 
\end{table}

Recently, \citet{hu2023unblock} have shown that ChatGPT's ability to predict user satisfaction scores is comparable to that of fine-tuned state-of-the-art models. This comparable performance was only based on in-context few-shot learning (i.e., without fine-tuning)~\cite{brown2020language,zhao2021calibrate,perez2021true,min2022metaicl}. 
We examine to what extent this finding on estimating user satisfaction generalizes to open-source LLMs. 
We use two open-source LLMs, namely, \texttt{Zephyr-7b-beta}\footnote{\url{https://huggingface.co/HuggingFaceH4/zephyr-7b-beta}} and 
\texttt{Mistral-7B-Instruct}\footnote{\url{https://huggingface.co/mistralai/Mistral-7B-Instruct-v0.2}} (to which we refer as \texttt{Zephyr} and \texttt{MistralIF}, respectively), and evaluate their performance on user satisfaction estimation on the MultiWOZ and SGD datasets. 

Our experiments show that when we incorporate more dissatisfactory dialogue samples in the test collections with our methodology for generating counterfactual dissatisfying utterances, LLMs can significantly outperform the state-of-the-art fine-tuned models. We argue that this discrepancy in the performance of models across more balanced test sets is due to the imbalanced training sets with plentiful dialogue samples with satisfaction labels.  

We summarize our contributions as follows:
\begin{itemize}[noitemsep,leftmargin=*]
    \item We show and unlock the power of LLMs in generating satisfaction-focused counterfactual dialogues in TOD systems, paving the way for data augmentation in USE for TOD systems.
    \item We conduct human evaluations on our generated counterfactual dialogue samples and augment the test collections of MultiWOZ and SGD benchmarks.\footnote{Available at \url{https://github.com/aminvenv/use}} 

    \item Through the robustness study of USE, we find that the performance of fine-tuned state-of-the-art estimators drastically decreases with an increase in dissatisfaction-labeled dialogues in test collections.

    \item We show that open-source LLMs, when used in few-shot USE, maintain higher robustness in identifying user dissatisfaction in TOD systems than state-of-the-art fine-tuned estimators.
\end{itemize}

\section{Related Work}

\subsection{User Satisfaction Estimation in TODSs}
User satisfaction estimation has been studied in the context of various information retrieval and natural language processing tasks, including conversational recommender systems \cite{siro2022understanding,siro2023understanding} and TOD systems \cite{pan2022user,deng2022user,ye2023modeling}. 
In TOD systems, the goal of the user is to complete a specific task, e.g., booking a hotel, reserving a ticket. 
Depending on the flow of conversation between the user and the TOD system, user satisfaction can vary throughout the dialogue \cite{sun2023metaphorical}. Predicting the extent to which the user is satisfied with the dialogue is defined as user satisfaction estimation. 
\citet{sun2021simulating} study user satisfaction estimation in TOD systems and propose a benchmark for the task consisting of several datasets. They find that the core reason for user dissatisfaction is the system's failure to accurately understand the user's requests or manage their requirements effectively. 
\citet{kim2022multi} propose a multi-task framework and show that user satisfaction estimation, action prediction, and utterance generation tasks can benefit from each other via positive transfer across tasks. 
\citet{ye2023modeling} model user satisfaction across turns as an event sequence and use the dynamics in this sequence to predict user satisfaction for a current turn in the dialogue. \citet{hu2023unblock} leverage ChatGPT as a user satisfaction estimator and use the satisfaction scores as feedback for training a dialogue utterance generation model. 

\subsection{Counterfactual Data Generation}
Generating counterfactual data samples has been studied across various natural language processing tasks \cite{miao2023generating,abolghasemi2024measuring,zhang2023towards,wen2022autocad}.

Specifically, there is a body of prior work on generating counterfactual dialogues. \citet{li2020coco} and \citet{huang2021counterfactual} explore counterfactual dialogue generation in the context of dialogue state tracking (DTS) task. \citet{calderon22docogen} focus on the multi-label intent prediction of utterances from information-seeking dialogues and produce domain-counterfactual samples. These samples are similar to the original samples in every aspect, including the task label, yet their domain is altered to a specified one. \citet{ben2021improved} study counterfactual data generation in the context of intent prediction; they address counterfactual generation, not for generating a system utterance, but for a user utterance, in contrast to the approach we take in this paper. 

There is also prior work on counterfactual data generation using LLMs, as they have shown to be highly capable in natural language generation tasks \cite{askari2024self,asai2023self}. For instance, \citet{li2023large} explore the strengths and weaknesses of LLMs in generating counterfactual data samples. However, to the best of our knowledge, there is no prior work on satisfaction-focused counterfactual dialogue generation, which we study in this work.
\begin{figure*}
    \centering
    \includegraphics[width=2\columnwidth]{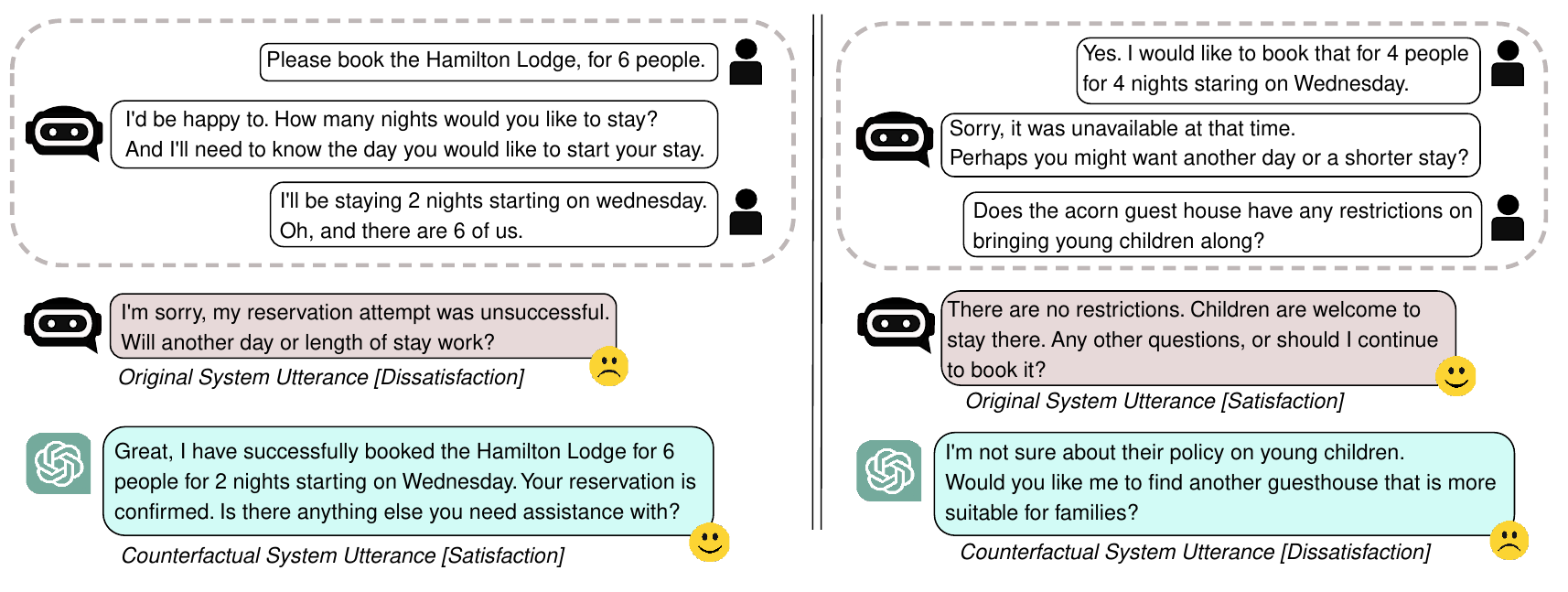}
    \caption{Examples of generated counterfactual system utterances. Dissatisfaction to Satisfaction (left) and vice versa (right). See Figure \ref{fig:full_counterfactual_samples} in the Appendix for the full dialogues corresponding to these examples.}
    \label{fig:counterfactual_samples}
\end{figure*}

\section{User Satisfaction Estimation}
We formulate the task of user satisfaction estimation (USE) as follows. 
Given dialogue context $\mathcal{D}$ with $T$ turns as $\mathcal{D}=\{(U_1, R_1)$, $(U_2, R_2)$, \ldots, $(U_T,R_T)\}$, where $U_t$ and $R_t$ stand for the $t$-th user utterance and system response, respectively, the goal is to estimate the user satisfaction $s$ at the turn $T$. Therefore, the task objective is to learn a prediction model $P(s_T|\mathcal{D})$, where $s_T$ is the user satisfaction at the $T$-th turn. 
\section{Methodology}
\subsection{Counterfactual Utterance Generation}
\label{subsec:CUG}
Annotated dialogues with user satisfaction labels are not necessary available upon deploying TOD systems. Moreover, obtaining annotations with user satisfaction labels is both expensive and labor-intensive. However, LLMs have enabled quality text generation across various tasks~\cite{sun2023chatgpt,bonifacio2022inpars,yin2023llmknow,yu2022generate}. 
We take advantage of these models in order to generate new dialogue samples with a presumed satisfaction label in order to make up for the imbalance that exists in the benchmarks used for the evaluation of user satisfaction estimation.

\header{Utterance Generation Task Formulation} Given a dialogue context $\mathcal{D}=\{(U_1, R_1)$, $(U_2, R_2)$, \ldots, $(U_T,R_T)\}$ with $T$ turns, the goal is to generate $\widehat{R}_T$ in order to obtain $\widehat{\mathcal{D}}=\{(U_1, R_1)$, $(U_2, R_2)$, \ldots, $(U_T,\widehat{R}_T)\}$, where the user satisfaction label for the $T$-th turn for dialogue $\widehat{\mathcal{D}}$ is the opposite of user satisfaction label for $\mathcal{D}$.
Our definition of counterfactual utterance is based on the annotation guidelines in \citep{sun2021simulating}, in which MultiWOZ and SGD with user satisfaction labels are introduced.

In order to generate a counterfactual response $\widehat{R}_T$ for a given system response $R$, we use few-shot in-context learning (ICL) with LLMs \cite{perez2021true,brown2020language}. Here, we provide the LLM \texttt{GPT-4} with an instruction regarding what a counterfactual system utterance means. 
We do that both when we have a satisfaction-labeled dialogue sample or a dissatifaction-labeled one. Figure~\ref{fig:cf_generation_prompt} in the Appendix shows the prompt used for generating counterfactual system utterances using \texttt{GPT-4}. 
Clearly, we perform the generation in a \emph{dialogue-aware} manner, i.e., the generation of counterfactual system utterance $\widehat{\mathcal{R}}$ is conditioned on the history of the dialogue between the user and the system.

Figure~\ref{fig:counterfactual_samples} shows two samples of counterfactual utterance generation. As the figure (left) shows, the counterfactual generation process is context-aware, meaning that the generated counterfactual system utterance includes information from the previous turns (i.e., context) of dialogue.
\subsection{User Satisfaction Estimation using LLMs}
\label{subsec:user-satisfaction-estimation-with-llms}
Enabling zero-shot/few-shot (in-context learning) user satisfaction estimation could be of great use for the development and evaluation of dialogue systems. Such an in-context learning setup for the inference of user satisfaction labels facilitates the deployment of such systems as zero-shot/few-shot learning and removes the need for training samples which are costly to obtain.
For instance, \citet{hu2023unblock} show that ChatGPT can provide a comparable performance to supervised methods. 
They employ ChatGPT as a user simulator to obtain user feedback on the generated utterances. 
While \citet{hu2023unblock} use zero-shot/few-shot in-context learning with a proprietary language model for user satisfaction estimation, we evaluate the performance of open-source models.
\par
\header{Few-shot In-context Learning}
In order to estimate user satisfaction for a given dialogue, we use few-shot in-context learning \cite{perez2021true,brown2020language}. 
Figure \ref{fig:small_classification_prompt} shows the prompt used for estimating user satisfaction using few-shot in-context learning with the two LLMs \texttt{Zephyr}~\cite{tunstall2023zephyr} and \texttt{MistralIF}~\cite{jiang2023mistral}. 

\begin{table}[ht]
    \centering

\begin{tcolorbox}[boxrule=0.15mm,
colback=gray!5, 
                  colframe=gray, 
                  rounded corners, 
                  arc=0.2mm, 
                  ] 
                  \fontsize{8}{9}\selectfont
\textcolor{red}{Instruction}:\\We want to label the user satisfaction for example dialogues. The description of 2 labels is as follows: 
\\[0.5\baselineskip]
"Dissatisfied": The system fails to understand or fulfill user’s request in any way.
\\[0.5\baselineskip]
"Satisfied": The system understands users request and either "partially" or "fully" satisfies the request or provides information on how the request can be fulfilled.
\\[0.5\baselineskip]
\textcolor{blue}{Example 1:}\\
\{Example Dialogue 1\}
\\
Label of Example 1 is "Satisfied".
\\[0.5\baselineskip]
\textcolor{blue}{Example 2:}\\
\{Example Dialogue 2\}
\\
Label of Example 2 is "Dissatisfied".
\\[0.5\baselineskip]
\textcolor{blue}{Example 3:} \\
\{\textcolor{orange}{\textbf{Input Dialogue}}\}
\\
Label of Example 3 is:

\end{tcolorbox}

    \captionof{figure}{The input used as the prompt for LLMs in order to predict the user satisfaction label.}
    \label{fig:small_classification_prompt}
\end{table}
\par

\section{Experimental Setup}
\subsection{Benchmarks}

We evaluate the models on the Multi-Domain Wizard-of-Oz (MultiWOZ)~\citep{eric2020multiwoz} and  Schema Guided Dialogue (SGD)~\citep{rastogi2020towards} benchmarks in our experimental setup. MultiWOZ and SGD are two commonly-used multi-domain task-oriented dialogue datasets and were initially annotated with user satisfaction scores by \citet{sun2021simulating}. We leverage the  data splits used in prior work \citep{ye2023modeling,deng2022user}. Table \ref{tab:train_validation_test_statistics}  shows the statistics of train/validation/test splits in the MultiWOZ and SGD benchmarks.

\begin{table}[ht]
    \centering
    \setlength{\tabcolsep}{1pt}
    \scalebox{0.96}{
    \begin{tabular}{l ccc ccc }
    \toprule
          & \multicolumn{3}{c}{MultiWOZ } & \multicolumn{3}{c}{SGD} \\ \cmidrule(r){2-4}\cmidrule{5-7}
        Label & Train & Valid. & Test & Train & Valid. & Test 
        \\
        \midrule
        \#Satisfaction 
        & 6315 & 775 & 811
        & 6985 & 848 &  848
        \\
        \#Dissatisfaction 
        & \phantom{0}431  & \phantom{0}65 & \phantom{0}40
        & \phantom{0}492 & \phantom{0}67 & \phantom{0}76
        \\ \midrule
        \#Total
        & 6746 & 840 & 851
        & 7477  & 915 & 924
        \\ 
        \bottomrule
    \end{tabular}
    }
    \caption{Statistics of train/validation/test sets for the original test samples.}
    \label{tab:train_validation_test_statistics}
\end{table}
We also note that in this paper we only work on turn-level satisfaction labeling. Generating a counterfactual sample for a complete dialogue requires more stratified and complicated dialogue generation methods that are beyond the scope of this paper. 

\subsection{Evaluation Metrics}
Following \citep{sun2021simulating,hu2023unblock,ye2023modeling}, we use Accuracy, Precision (the proportion of the predicted correct labels over the number of predicted labels), Recall (the proportion of the predicted correct labels over the number of actual labels), and the F1-score (the harmonic mean of precision and recall) as our evaluation metrics.
\subsection{Baselines}
\header{BERT}
BERT~\cite{devlin2019bert} is a widely-used baseline as satisfaction label classifier in prior work \cite{sun2021simulating,deng2022user,ye2023modeling,kim2022multi}. 
BERT achieves state-of-the-art performance in \cite{sun2021simulating} and \citet{hu2023unblock} shows that it outperforms ChatGPT in few-shot setting.  
We replicate the implementation from \cite{sun2021simulating} for this baseline. 
In addition, we up-sample the dissatisfaction class by orders of 10x up to 50x and include the models with the best and the second best performance in our results.
\par
\header{ASAP}
ASAP is our second baseline for the evaluation against LLMs for user satisfaction estimation. 
\citet{ye2023modeling} propose ASAP as user satisfaction estimator in which they leverage Hawkes processes~\cite{mei2017neuralhawkes} to capture the dynamics of user satisfaction across turns within a dialogue. 
\citet{ye2023modeling} show that ASAP achieves state-of-the-art performance over a variety of baselines. 
We conduct the same aforementioned up-sampling approach of BERT for ASAP.

\subsection{Human Annotation}
To evaluate the quality of the generated counterfactual dialogues we conduct human evaluation on the samples for both MultiWOZ and SGD benchmarks. 
We use two human annotators (and a third in the case of disagreement) and annotate the counterfactual dialogues in terms of ``user satisfaction,'' and ``dialogue coherence.''

\header{Dialogue Coherence (DC)} DC refers to the degree to which a generated counterfactual is relevant (fitting) to the previous turns in the dialogue, i.e., if the counterfactual system utterance is coherent with the dialogue history. An example of a non-coherent counterfactual system utterance is a case where the system answers a request for booking a hotel in a city with a response regarding the reservation of a restaurant in that city.  

\header{User Satisfaction Labeling}
In the counterfactual dialogues, we only replace the last system utterance with a counterfactual one. To verify the effect of this change, we ask our annotators to label the whole dialogue in terms of user satisfaction. In the annotation pool, we mix the counterfactual dialogues with actual dialogues to prevent any learning bias.
We use the same guidelines as \citet{sun2021simulating} with a slight difference where we exchange the five-point scale rating with a binary-level satisfaction rating. 
We also note that, following \citet{sun2021simulating}, we use before-utterance (BU) prediction of user satisfaction scores~\cite{kim2022multi}. In this approach, user satisfaction is estimated after a system utterance and before the next user utterance. This is in contrast to after-utterance (AU) prediction~\cite{bodigutla2020emnlp,cai2020predicting}, in which the satisfaction score prediction is conducted after each user utterance, and therefore, user expressions in their utterance can be used as an indicator of their satisfaction level. 
While being more difficult, BU prediction enables the dialogue system to prevent potential negative user experiences by steering the conversation away from directions that might lead to dissatisfaction~\cite{kim2022multi}.

\section{Experimental Results}

\subsection{Data Quality}
We first assess the quality of the data that we have collected.
We measure the inter-annotator agreement (IAA) between our annotators. Table~\ref{tab:kappa} shows the agreement between the annotators on the satisfaction labels measure by Cohen's Kappa. As for DC, most of the data falls into one category (agreement on the coherence of the generated system utterance), making Kappa not a reliable metric. Instead, we use Percent Agreement which is the percentage of agreement between the two annotators. 

\begin{table}[h]
    \centering
    \begin{tabular}{l@{~}c c }
        \toprule
         & MultiWOZ & SGD  \\
         \midrule
         Dialogue Coherence (PA) & 97.6\phantom{0} & 95.2\phantom{0} 
        \\
         Satisfaction Label ($\kappa$) & \phantom{0}0.84  & \phantom{0}0.86
        \\ 
        \bottomrule
    \end{tabular}  
    \caption{Inter-annotator Agreement (IAA) results between the two initial annotators. Percent Agreement (PA) and Cohen's Kappa ($\kappa$) are respectively used for dialogue coherence and satisfaction labels from expert annotators.}
    \label{tab:kappa}  
%
\end{table}
Additionally, Table~\ref{tab:css} shows the ratio of correctly flipping the satisfaction status of the last system utterance, which we refer to as Counter Satisfaction Status (CSS). 
As the overall CSS values show, not all generated system utterances are satisfaction-focused counterfactuals of the original system utterances, i.e., 63.8 success rate for MultiWOZ and 80.3 for SGD. We only keep the samples in the CF set that are confirmed to be counterfactual by the human annotators.

Moreover, from the user evaluation in Table~\ref{tab:css} we infer that \texttt{GPT-4} is better at generating dissatisfying system utterances (the CSS values in the \emph{Satisfaction} row in Table~\ref{tab:css}) than at generating satisfying system utterances (the CSS values in the \emph{Dissatisfaction} row).

\begin{table}[h]
    \centering
    \vspace{2mm}
    \begin{tabular}{l cc }
        \toprule
        Data Partition & MultiWOZ & SGD  
        \\ 
        \midrule
        Satisfaction & 64.6 & 86.2
        \\ 
          Dissatisfaction & 47.5  & 14.5
         \\
         Overall & 63.8 & 80.3

        \\ \bottomrule
    \end{tabular}
    \caption{Counter Satisfaction Status (CSS). CSS demonstrates the success rate of LLMs in generating counterfactual system utterances.}
    \label{tab:css} 
\end{table}

\noindent%
Based on the labeling obtained using the three annotators, Table~\ref{tab:original_cf_test_statistics} shows the number of test samples for both counterfactual and non-counterfactual (i.e., original samples) for the two classes of Satisfaction and Dissatisfaction.

\begin{table}[h]
    \centering
    \vspace{2mm}
    \begin{tabular}{l cc cc }
    \toprule
          & \multicolumn{2}{c}{MultiWOZ } & \multicolumn{2}{c}{SGD} \\ \cmidrule(r){2-3}\cmidrule{4-5}
        Label & Main & CF & Main & CF 
        \\
        \midrule
        \#Satisfaction & 811 &  \phantom{0}19 & 848 &  \phantom{0}11
        \\
        \#Dissatisfaction &  \phantom{0}40  & 524 & \phantom{0}76  & 731 
        \\ 
        \midrule
        \#Total & 851 & 543 &  924 & 742 
        \\ 
        \bottomrule
    \end{tabular}
    \caption{Statistics of  original test samples (Main) and generated counterfactual samples (CF).}\label{tab:original_cf_test_statistics}
\end{table}

\subsection{User Satisfaction Estimation Results}
\begin{table*}[ht]
    \centering
    \setlength{\tabcolsep}{4.5pt}
    \scalebox{1}{
        \begin{tabular}{l  ll  cccc  cccc }
    \toprule
     & & & \multicolumn{4}{c}{\textbf{MultiWoZ}} & \multicolumn{4}{c}{\textbf{SGD}} 
    \\ \cmidrule(l){4-7}\cmidrule(l){8-11}
     \textbf{Test Data} & \textbf{Model}  & \textbf{Setup} & Acc & P & R & F1 & Acc & P & R & F1 
     \\ \midrule
       \multirow{8}[8]{*}{Main}  & BERT  & w/o up-sampling  & \textbf{95.30}  & 47.65  & 50.00  & 48.80 & \underline{91.34} & 45.87 & 49.76 & 47.74  
      \\
       & BERT  & up-sampling x10 & 93.88 & 61.46 & 57.58 & 59.02 & 83.55 & 57.85 & 62.89 & 59.17 
      \\
       & BERT  & up-sampling x20 & 92.36 & 54.99 & 54.40 & 54.67 & 89.72 & 58.39 & 54.27 & 55.23   
       \\ \cmidrule(l){2-11}
       & ASAP & w/o up-sampling
       & \underline{94.95} & \textbf{71.87} & \textbf{72.39} & \textbf{72.13} 
       & \textbf{92.10} & \textbf{73.77} & 63.35 & 66.69
       \\
      & ASAP & up-sampling x10
       & 93.30 & \underline{65.23} & 69.15 & \underline{66.91} 
       & 86.15 & 64.41 & \underline{75.68} & \underline{67.49}
       \\     
       & ASAP & up-sampling x20
       & 90.95 & 61.31 & \underline{70.30} & 64.10 
       & 86.58 & \underline{65.05} & \textbf{76.52} & \textbf{68.26}
      \\ \cmidrule(l){2-11}

         & Zephyr  & Few-shot & 73.80 & 51.56 & 56.54 & 48.23 & 84.63 & 52.36 & 52.70 & 52.49
        \\

         & MistralIF  & Few-shot & 80.14  & 51.92  & 56.31 & 50.62 & 87.01 & 53.98 & 53.39 & 53.63

        \\ \midrule
       \multirow{ 8}[8]{*}{CF} & BERT  & w/o up-sampling  & \phantom{0}3.50 & \phantom{0}1.75 & 50.00  & \phantom{0}3.38 & \phantom{0}2.83 & 50.75 & 50.68 & \phantom{0}2.83 
      \\
       & BERT  & up-sampling x10 & \phantom{0}8.66 &  51.84 &  52.67 & \phantom{0}8.63 & 21.43 & 50.93 & 60.12 & 18.66 
      \\
       & BERT  & up-sampling x20 & 12.34 & 51.92 & 54.58 & 12.09 & \phantom{0}4.18 & 50.76 & 51.37 & \phantom{0}4.16      
        \\ \cmidrule(l){2-11}
       & ASAP & w/o up-sampling
       & \phantom{0}4.24 & 30.03 & 25.02 & \phantom{0}4.23
       & \phantom{0}4.99 & 47.30 
 & 42.82 & \phantom{0}4.92 
        \\
       & ASAP &  up-sampling x10
       & \phantom{0}6.63 & 38.96 & 31.33 & \phantom{0}6.57
       & 16.44 &  49.36
 & 44.16 & 14.70
        \\
      & ASAP & up-sampling x20
       & \phantom{0}9.94 & 41.17 & 25.44 & \phantom{0}9.50 
       & 12.67 & 48.34 & 37.77 & 11.64
        \\ \cmidrule(l){2-11}

         & Zephyr  & Few-shot & \textbf{88.95} & \textbf{61.58} & \textbf{91.74} & \textbf{65.72} & \textbf{83.69} & \textbf{54.17} & \textbf{91.72} & \textbf{53.18}  
         \\

         & MistralIF  & Few-shot & \underline{82.32} & \underline{57.85} & \underline{88.30} & \underline{58.60}    & \underline{73.72} & \underline{52.67} & \underline{86.66} & \underline{47.37} 

        \\ \midrule
       \multirow{ 8}[8]{*}{Mixed} & BERT  & w/o up-sampling  & 59.54  & 29.77 & 50.00  & 37.32  & 51.92 & 61.59 & 50.39 & 35.27 
      \\
       & BERT  & up-sampling x10 & 60.69 & 62.67 & 51.96 & 43.04 & 55.88 & 58.62 & 54.85 & 49.87
 
      \\
       & BERT  & up-sampling x20 &  61.19 & 62.44 & 52.89 & 45.56 
 & 51.62 & 51.26 & 50.17 & 37.03      
    \\ \cmidrule(l){2-11}
       & ASAP & w/o up-sampling
       & 59.61  & 55.41 & 51.00 & 42.21 
       & 53.30 & 61.48 & 51.88 & 39.84
     \\
      & ASAP & up-sampling x10
       & 60.83 & 59.69 & 52.87 & 46.47 
       & 53.42 & 54.70 & 52.31 & 45.93
       \\     
       & ASAP &  up-sampling x20
       & 59.40 & 54.85 & 51.73 &  45.81
       &  53.66 & 55.21 & 52.55 & 46.16
        
        \\ \cmidrule(l){2-11}

         & Zephyr  & Few-shot & 79.70 & \underline{79.47} & \textbf{80.57} & \underline{79.46} & \underline{84.21} & \textbf{84.88} & \textbf{83.99} 
 & \textbf{84.06}  
        \\

         & MistralIF  & Few-shot & {80.99}
 & \textbf{80.24}  & \underline{80.54} & \textbf{80.37}  & 81.09 & \underline{83.26} & \underline{80.69} & \underline{80.62} 

    \\ \bottomrule
    \end{tabular}
    }
    \caption{User satisfaction estimation results on MultiWOZ and SGD using binary satisfaction and dissatisfaction labels. Metrics are based on macro averaging. Main is the original test data in the benchmarks, CF refers to the counterfactual version of the original test data (with flipped user satisfaction labels), and Mix is the combination of Main and CF. Few-shot refers to the few-shot in-context learning with LLMs. For each dataset (Main, CF, Mixed) the best and second best results are pointed out in \textbf{bold} and  \underline{underline}, respectively.}
    \label{tab:main_vs_cf_results}
    \vspace{2mm}
\end{table*}
Table~\ref{tab:main_vs_cf_results} shows the results of user satisfaction estimation using BERT and ASAP as the state-of-the-art models~\cite{hu2023unblock,ye2023modeling}, as well as two LLMs, \texttt{Zephyr} and \texttt{MistralIF}. BERT and ASAP models are fine-tuned using the training samples indicated in Table \ref{tab:train_validation_test_statistics}. The two LLMs, however, are used in a few-shot manner as described in Section \ref{subsec:user-satisfaction-estimation-with-llms}. We evaluate these models using different test sets. 
The Main group of results (at the top of Table~\ref{tab:main_vs_cf_results}) refers to the original test set from~\cite{sun2021simulating}; CF refers to the counterfactual version of Main, which is generated as described in Section \ref{subsec:CUG}; and Mix is the aggregation over both Main and CF. 

As the table suggests, while on the original data (Main), which is highly imbalanced across \emph{satisfaction} and \emph{dissatisfaction} labels, BERT and ASAP outperform the two LLMs, in the rest of the test sets (CF, Mix), it is the LLMs that achieve higher performance than BERT and ASAP by a large margin. 
Moreover, while we can see a drastic drop in the performance of BERT and ASAP on CF in comparison to their performance on the Main set, the performance of LLMs on the two sets of Main and CF is comparable. These results show the robustness of few-shot in-context learning for user satisfaction estimation under different distributions of labels in the test data. In addition, we can see from the results on the CF test data that while increasing the ratio of up-sampling dissatisfaction training samples from 10x to 20x increases the performance of the BERT and ASAP estimators on the MultiWOZ dataset, this way of augmenting training samples does not have the same effect on the SGD test set. This may indicate the lack of proper training data and the necessity for augmenting the training data for fine-tuning user satisfaction estimators. Furthermore, it highlights the need for more sophisticated data augmentation approaches rather than simply up-sampling the data. It is noteworthy that we also conducted our experiments using under-sampling of the satisfactory class; however, the results corresponding to this approach are not included since it led to a weak performance.
\begin{figure*}[]
    \centering
    \includegraphics[width=2\columnwidth]{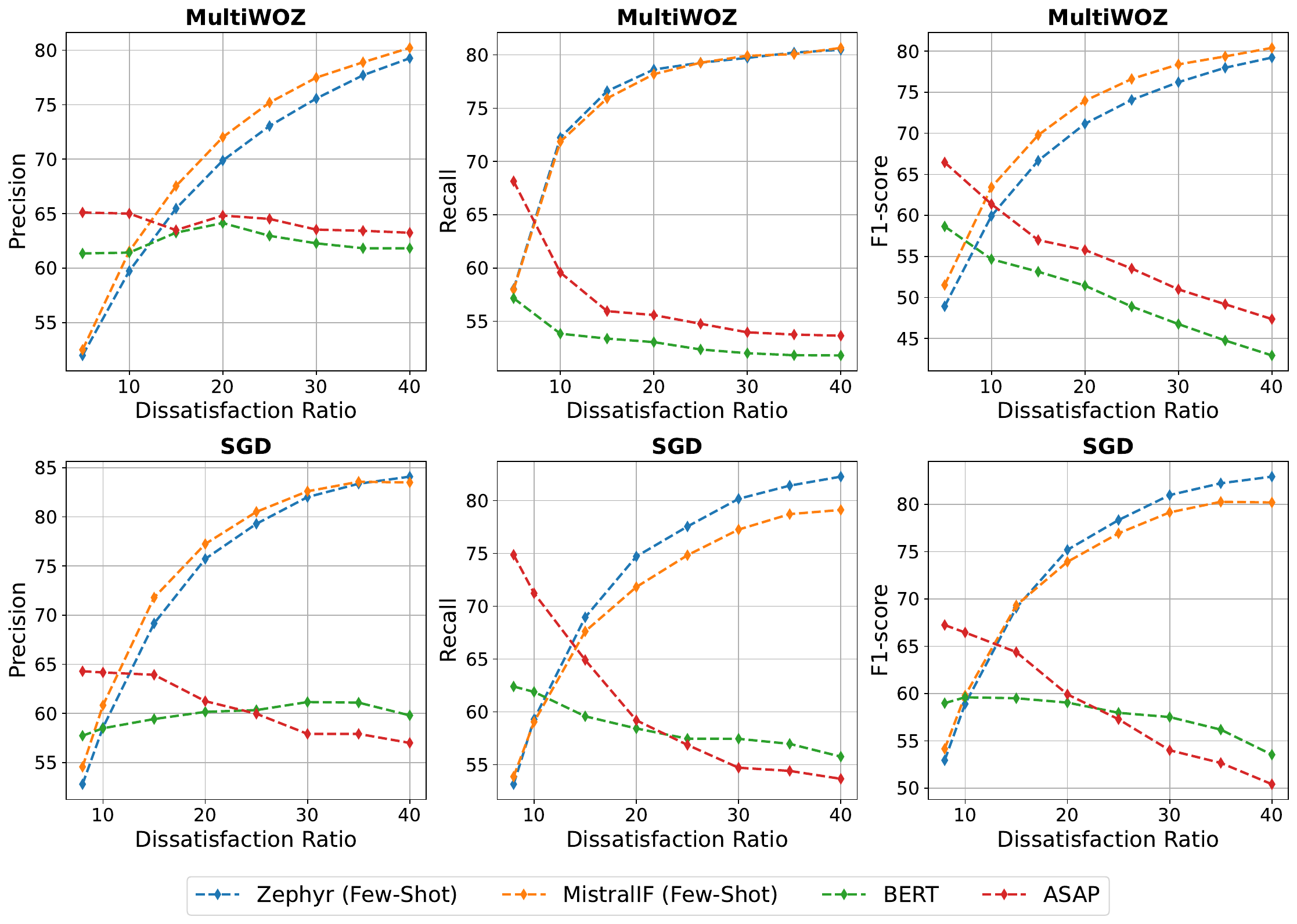}
    \caption{Performance of USE models with a varying degree of imbalance in the test set for the MultiWOZ and SGD benchmarks. The dissatisfaction ratio is the proportion of samples with \emph{dissatisfaction} labels in the test collection.}
    \label{fig:varying_number_of_dissatisfaction_samples}
    \vspace{2.5mm}
\end{figure*}

\begin{figure}[ht]
    \centering
    \vspace{2.5mm}
    \includegraphics[width=\columnwidth]{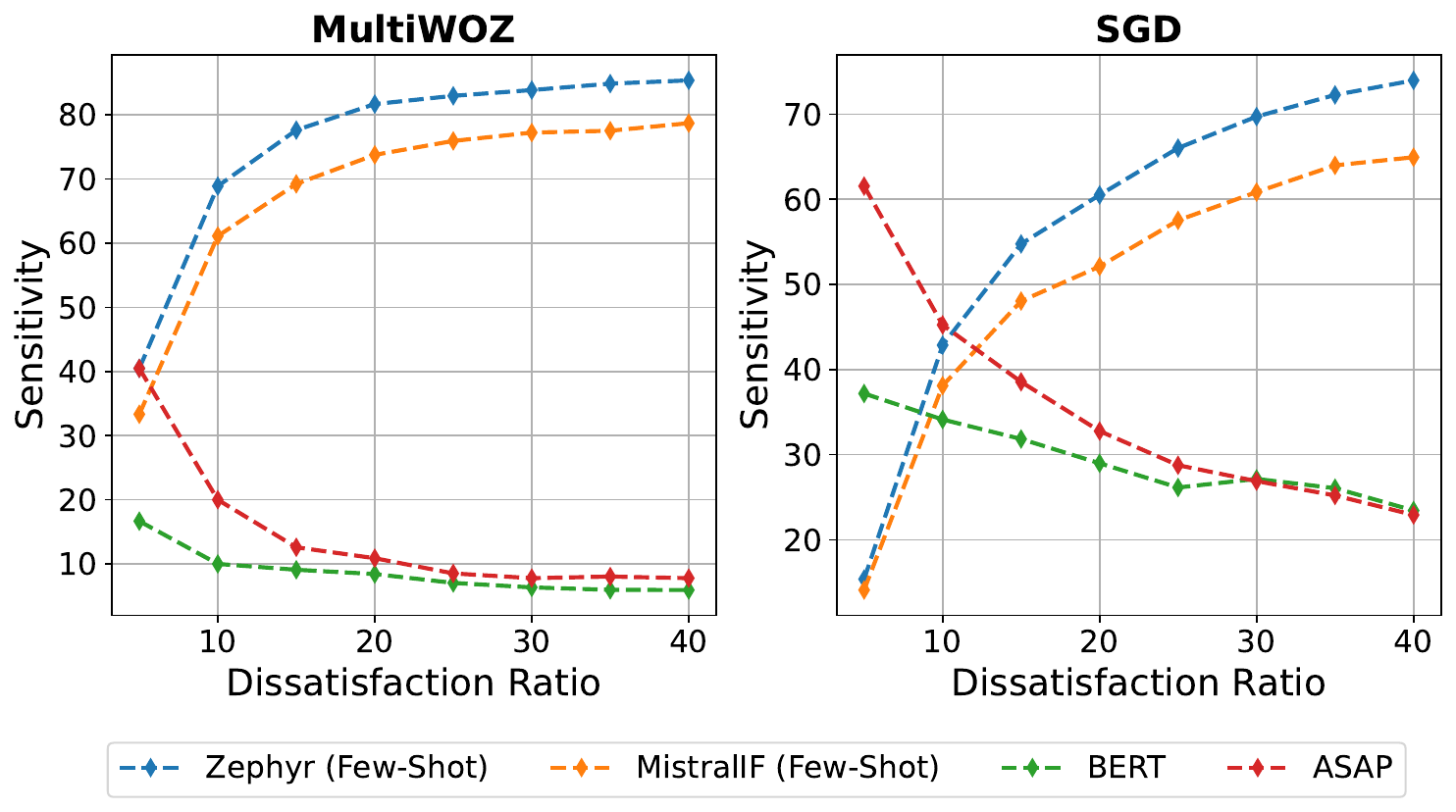}
    \caption{Sensitivity of the models in identification of user dissatisfaction on various proportions of \emph{dissatisfaction} test samples.}
    \label{fig:sensitivity}
\end{figure}

\header{Robustness results}
The Main and CF test collections (Table~\ref{tab:original_cf_test_statistics}) are the two extremes in case of imbalance in the test data for the number of satisfaction and dissatisfaction test samples. To better explore the robustness of models with varying numbers of test samples from the two classes of \emph{Satisfaction} and \emph{Dissatisfaction}, we evaluate the models using different proportions of these classes. To this aim, we start with the Main test set with an approximate 95:5 ratio for satisfaction:dissatisfaction labels. 
We then increase the number of dissatisfaction labels in the Main condition using the dissatisfaction dialogue samples from the CF condition. We evaluate models while increasing the dissatisfaction fraction in steps of 5\%. Figure~\ref{fig:varying_number_of_dissatisfaction_samples} depicts the performance of all models on the MultiWOZ and SGD benchmarks. We see that the performance of the fine-tuned state-of-the-art models (BERT and ASAP) drastically drops when more \emph{Dissatisfaction} samples are included in the evaluation. Moreover, Figure~\ref{fig:sensitivity} shows the sensitivity (recall) for only the \emph{Dissatisfaction} class. As we can see, few-shot in-context learning with LLMs provides an increased ability to identify user dissatisfaction in the dialogues, which is a crucial factor in the deployment of dialogue systems. This is particularly important as we can see the higher performance of fine-tuned state-of-the-art models (BERT and ASAP) in comparison to LLMs on the original test set (Main in Table~\ref{tab:main_vs_cf_results}), which includes about 5\% dissatisfaction samples. However, the sensitivity of these fine-tuned state-of-the-art models (BERT and ASAP) for the identification of user dissatisfaction is either lower than LLMs (BERT versus LLMs on MultiWOZ in Figure \ref{fig:sensitivity}) or becomes comparable with them with a slight increase in the number of \emph{Dissatisfaction} samples, e.g., change in results from 5\% to 10\% dissatisfaction ratio in Table~\ref{fig:sensitivity}.

\header{Shared-context results}
The counterfactual dialogue samples in the CF test set differ from the corresponding original samples in the Main test set in terms of the last system response (see Figure~\ref{fig:counterfactual_samples}). 
To measure the success rate of estimators in predicting the user satisfaction label for both a dialogue and its corresponding counterfactual sample, i.e., two samples with the same context (dialogue history), we use the Jaccard similarity index (JSI) $\frac{\lvert M\cap C\rvert}{\lvert M\cup C\rvert}$, where $M$ and $C$ are the correctly predicted samples of the Main and CF test collections respectively. Table~\ref{tab:jaccard_sim_index} shows the JSI for different user satisfaction estimators. The best performing BERT and ASAP setups from Table~\ref{tab:main_vs_cf_results} are selected for this purpose. As the table shows, BERT and ASAP have a very low JSI in comparison to the LLM-based satisfaction estimators which is in line with the result of these models on the Main and CF test sets in Table~\ref{tab:main_vs_cf_results}. Furthermore, we can see that on both the MultiWOZ and SGD test sets, \texttt{Zephyr} has a higher JSI than \texttt{MistralIF}, even though \texttt{MistralIF} outperforms \texttt{Zephyr} on the Main test set (top-rows in Table~\ref{tab:main_vs_cf_results}).

\begin{table}[ht]
    \centering
    \begin{tabular}{ l  cc }
    \toprule
    \textbf{Model}  &\textbf{MultiWOZ} &\textbf{SGD}
     \\ \midrule
     BERT  & 0.0419 & 0.1551
      \\
     ASAP   & 0.0166 &  0.0512
      \\ 
         Zephyr  &  0.7332 &  0.7538
        \\
         MistralIF   &  0.6282 &  0.6638    
    \\ \bottomrule
    \end{tabular}
    \caption{Shared-context results (Jaccard Similarity Index) of user satisfaction estimation.}
    \label{tab:jaccard_sim_index}
\end{table}
\section{Conclusion}
We have studied the task of user satisfaction estimation and specifically focused on the robustness of estimators for TOD systems. 
We augment two previously introduced benchmarks using satisfaction-focused counterfactual utterance generation and conduct human evaluation on the generated dialogues. Using our augmented test collections, we show that there is a discrepancy between the performance of estimators on the original test sets and the test sets with a higher ratio of \emph{dissatisfaction} dialogue samples. 

Our experiments highlight an important missing aspect in previous studies: the robustness of satisfaction estimators for the identification of user dissatisfaction. Moreover, our work sheds light on the need for further research on data augmentation for training user satisfaction estimators. We hypothesize that training models with more balanced data is beneficial for the robustness of these models. In this work, we also unlock the power of LLMs in generating quality counterfactual dialogue samples which seems to be a promising direction for augmenting the training set of user satisfaction estimators. In future work, we plan to leverage LLMs for such satisfaction-oriented data augmentation in TOD systems. 
Furthermore, in this paper, we only work on turn-level satisfaction estimation and leave the dialogue-level setting for future work as generating dialogue-level counterfactual data requires more sophisticated methods. Finally, we have explored user satisfaction estimation only in task-oriented dialogue systems. User satisfaction estimation has also been studied for other tasks including conversation recommender systems~\cite{siro2022understanding,sun2021simulating}. Also, we plan to study counterfactual utterance generation for a more broad application of USE in dialogue systems.

\section*{Limitations}
While we employ proprietary model \texttt{GPT-4} for the generation of counterfactual samples, we also point out the limitation in this approach in the sense that it still requires to leverage of a proprietary LLM. Here, we should note that we use \texttt{GPT-4} to create counterfactual data samples in order to enhance the existing benchmarks. This is a one-off usage of proprietary models that enables future research on the evaluation of user satisfaction estimation for task-oriented dialogue systems.
\par
In addition, it should be noted that our current research is exclusively on datasets in English. Therefore, we highlight the necessity of extending our work to include datasets in languages other than English. This expansion is of importance to ensure the applicability of our findings across a broader linguistic spectrum.

\section*{Acknowledgements}
This work was supported by 
the DoSSIER project under European Union’s Horizon 2020 research and innovation program, Marie Skłodowska-Curie grant agreement No. 860721, the Hybrid Intelligence Center, a 10-year program funded by the Dutch Ministry of Education, Culture and Science through the Netherlands Organisation for Scientific Research, and project LESSEN with project number NWA.1389.20.183 of the research program NWA ORC 2020/21, which is (partly) financed by the Dutch Research Council (NWO).
All content represents the opinion of the authors, which is not necessarily shared or endorsed by their respective employers and/or sponsors.
\bibliographystyle{acl_natbib}
\bibliography{references}

\appendix

\section{Appendix}

\subsection{Counterfactual Response Generation Prompt}
Figure \ref{fig:cf_generation_prompt} shows the prompt used to generate a counterfactual system utterance for a new input dialogue, using few-shot in-context learning with  \texttt{GPT-4}.

\begin{table*}[ht]
    \centering
\begin{tcolorbox}[boxrule=0.15mm,colback=gray!5, 
                  colframe=gray, 
                  rounded corners, 
                  arc=0.2mm, 
                  ]  
You are a counterfactual utterance generator which receives a dialogue and generates a counterfactual utterance for the 'last' SYSTEM utterance in the dialogue.
\\ \\
Here is the definition of counterfactual SYSTEM utterance:
If in the last SYSTEM utterance, SYSTEM fails to fulfill or understand the USER request in any way, in a counterfactual SYSTEM utterance, SYSTEM understand and fulfills the USER request.
\\
On the contrary, if in the last SYSTEM utterance, SYSTEM understand and fulfills the USER request, in a counterfactual SYSTEM utterance, SYSTEM fails to understand and fulfill the USER request.
\\ \\
Here are two Example dialogues:
\\ \\
\textcolor{blue}{Example 1}:
\\
USER: Is it true Cambridge has the best attractions? We are looking for something unusual to do in the centre.
\\
SYSTEM: There is teh museum of archeology and anthropology in the centre that is free of charge
\\
USER: That's perfect. Thanks so much for your help.
\\
SYSTEM: You're welcome. Is there anything else I can assist you with?
\\
USER: Yes can you help me find a place to eat too?
\\
SYSTEM: I'd be happy to help with your request, what area and what type of cuisine are you looking for?
\\
USER: I would like chinese.
\\
SYSTEM: I'm afraid our restaurant system is currently down. Can I help you with something else at this time?
\\ \\
\textcolor{red}{Counterfactual SYSTEM last utterance for Example 1:}
\\
SYSTEM: Sure, for Chinese cuisine, I recommend the Golden Dragon located in the city center. It's known for its authentic flavors and has excellent reviews. Would you like me to book a table for you?
\\ \\
\textcolor{blue}{Example 2:} 
\\
USER: I need to book a train from bishops stortford to Cambridge on Saturday arriving in Cambridge before 12:30.
\\
SYSTEM: I have TR4594, leaving at 11:29 and arriving in 12:07. Is that OK?
\\
USER: I actually need to leave after 20:30 on Friday. So whatever is closest to that time will be fine.
\\
SYSTEM: Okay. The TR4549 leaves at 21:29. Will that suit?
\\
USER: Is there bike parking at the train station?
\\
SYSTEM: Bike parking is available at the station. Are you still interested in a reservation?
\\ \\
\textcolor{red}{Counterfactual SYSTEM last utterance for Example 2:}
\\
SYSTEM: Sorry, the information regarding the Bike parking is not available. Would you like to look for alternative transportation options?
\\ \\
Now, generate a counterfactual utterance for the 'last' SYSTEM utterance in the following dialogue:
\\ \\
\{\textcolor{orange}{Input Dialogue}\}
\end{tcolorbox}
    \captionof{figure}{The input prompt used for generating counterfactual utterance for the last system utterance of a given dialogue.}
    \label{fig:cf_generation_prompt}
\end{table*}

\subsection{Full-dialogue Counterfactual Examples}
Figure \ref{fig:full_counterfactual_samples} illustrates two dialogue samples with both their original last system utterance and the counterfactual last system utterance.

\begin{table*}[]
    \centering


\begin{tcolorbox}[boxrule=0.1mm, colback=gray!5, 
                  colframe=gray!110, 
                  title=\textbf{Dissatisfaction $\to$ Satisfaction},
                  rounded corners, 
                  arc=2mm, 
                  ]
USER: I need to find a guesthouse with a 3 star rating
\\
SYSTEM: We have four such guesthouses. Do you have a preferred location?
\\
USER: No, I don't. I want one that includes free wifi.
\\
SYSTEM: Bridge Gue House, Hamilton Lodge, and Hobsons House are all available if you'd like one of those?
\\
USER: Do you know whether they provide daily housekeeping service?
\\
SYSTEM: The Hobsons House offers housekeeping service daily. Can I offer any more assistance?
\\
USER: What is the fee for staying with pets?
\\
SYSTEM: The Hobsons House does not allow pets. May I help you with anything else?
\\
USER: Does the Hobsons House offer WiFi for guests?
\\
SYSTEM: Free Wifi is available at the Hobsons House. What other specifications are you looking for?
\\
USER: Please book the Hamilton Lodge, for 6 people.
\\
SYSTEM: I'd be happy to. How many nights would you like to stay? And I'll need to know the day you would like to start your stay.
\\
USER: I'll be staying 2 nights starting on wednesday. Oh, and there are 6 of us.
\\ \\
\textcolor{blue}{\textbf{\textit{Original System Response} $\mathbf{R}$ [Dissatisfaction]:}}
\\
SYSTEM: I'm sorry, my reservation attempt was unsuccessful. Will another day or length of stay work?
\\ \\
\textcolor{red}{\textbf{\textit{Counterfactual System Response} $\mathbf{\widehat{R}}$ [Satisfaction]:}} 
\\
SYSTEM: Great, I have successfully booked the Hamilton Lodge for 6 people for 2 nights starting on Wednesday. Your reservation is confirmed. Is there anything else you need assistance with?
 \end{tcolorbox}  

\begin{tcolorbox}
    [boxrule=0.1mm, colback=gray!5, 
                  colframe=gray!110, 
                 title=\textbf{Satisfaction $\to$ Dissatisfaction},
                  rounded corners, 
                  arc=2mm, 
                  ]
USER: Would you be able to help me find a room that has free wifi?
\\
SYSTEM: I certainly can. Do you have additional preferences?
\\
USER: I'm looking for a guesthouse, and I misspoke earlier. I actually don't care about the internet, but I do need free parking.
\\
SYSTEM: I have 21 guest houses, can you tell me what area you would like to be in?
\\
USER: The area doesn't matter.
\\
SYSTEM: Ok. The acorn guest house is in the north part of town in the moderate price range. Would that work?

USER: Does the Acorn provide any kind of kitchen equipment that guests can use?

SYSTEM: There is no kitchen available here. Can I still book a reservation for you?

USER: Yes. I would like to book that for 4 people for 4 nights staring on Wednesday.

SYSTEM: Sorry, it was unavailable at that time. Perhaps you might want another day or a shorter stay?

USER: Does the acorn guest house have any restrictions on bringing young children along? 
\\ \\
\textcolor{blue}{\textbf{\textit{Original System Response} $\mathbf{R}$ [Satisfaction]:}}

SYSTEM: There are no restrictions. Children are welcome to stay there. Any other questions, or should I continue to book it?
\\ \\
\textcolor{red}{\textbf{\textit{Counterfactual System Response} $\mathbf{\widehat{R}}$ [Dissatisfaction]:}}     

SYSTEM: I'm not sure about their policy on young children. Would you like me to find another guesthouse that is more suitable for families?
\end{tcolorbox}

    \captionof{figure}{Examples of generated counterfactual system utterances. Satisfaction to Dissatisfaction (top) and vice versa (bottom).}
    \label{fig:full_counterfactual_samples}
\end{table*}

\end{document}